# On filter design in deep convolutional neural network

Gaurav Hirani[1*] and Waleed Abdulla[2]

[12]The University of Auckland, Auckland, New Zealand

[1]E-mail: ghir325@aucklanduni.ac.nz    Tel: +64 22 307 2932

[2]E-mail: w.abdulla@auckland.ac.nz    Tel: +64 (9) 373 7599 Ext.88969

**Abstract:**
The deep convolutional neural network (DCNN) in computer vision has given promising results. It is widely applied in many areas, from medicine, agriculture, self-driving car, biometric system, and almost all computer vision-based applications. Filters or weights are the critical elements responsible for learning in DCNN. Backpropagation has been the primary learning algorithm for DCNN and provides promising results, but the size and numbers of the filters remain hyper-parameters. Various studies have been done in the last decade on semi-supervised, self-supervised, and unsupervised methods and their properties. The effects of filter initialization, size-shape selection, and the number of filters on learning and optimization have not been investigated in a separate publication to collate all the options. Such attributes are often treated as hyper-parameters and lack mathematical understanding. Computer vision algorithms have many limitations in real-life applications and understanding the learning process is essential to have some significant improvement. To the best of our knowledge, no separate investigation has been published discussing the filters; this is our primary motivation. This study focuses on arguments for choosing specific physical parameters of filters, initialization, and learning technic over scattered methods. The promising unsupervised approaches have been evaluated. Additionally, the limitations, current challenges, and future scope have been discussed in this paper.

## 1 | INTRODUCTION

The convolution neural Network (CNN) has become the pioneer method for applying the neural network concept to computer vision tasks [1]. The advancement of the basic CNN-LeNet-5 (5 layers) model to more profound and complex models like AlexNet (8 layers) [2], VGG (11-19 layers) [3], ResNet (152 layers) [4], GoogleNet (22 layers) [5] have achieved superior efficiency in real-life applications. Such models are called deep CNN (DCNN), the combinations of deep learning structure and CNN. Even though DCNN is based on specific mathematical models, due to its more profound and complex design and nonlinear activation functions, the fundamental insight remains as a Blackbox [6].

The fundamental constituents of DCNN are filters, activation functions, and classifiers. The structure is divided into two major sections: feature extraction and classification. Filters are often known as weights that do the actual magic when it comes to learning, and it is very critical to understand the exact end-to-end processing in them. In all the layers of DCNN, different filters perform convolution operations with the input to that layer for feature extraction. Activation functions control the dynamics of information flow from one layer to the next. A standard ML algorithm classifier categorizes the classes based on learned features through convolution. It is more suitable to say that the primary design attributes of filters are unknown, along with chosen training algorithm and activation functions. The interrelation of filter hyper-parameters, training algorithms, and activation functions lack concrete theoretical background for DCNN.

1.1 | Convolutional neural network (CNN)

The convolutional neural Network (CNN) has dominated computer vision for almost three decades. To briefly introduce, there can be nothing better than LeNet-5, a pioneering architecture given in 1998 [1]. LeNet-5 is made of a total of 7 layers: 3 convolutional layers (C1, C3, and C5), and two subsampling layers (S2 and S4), followed by two fully connected layers (F6 and F7). The core section is the convolutional layers, where the actual learning happens. Images have a 2D spatial structure and can have spectral dimensions for color images (e.g., 3-Dimentional (3D) for RGB images). Convolution can happen to the whole image at once, which would give a scalar point and is not helpful for multidimensional feature learning. Hence input image is divided into smaller image patches, often called activation space. The filter(s) convolve with activation space sliding over the input image in an overlapping or no-overlapping fashion and giving a scalar point for each activation space. The output goes through an activation function, e.g., Rectified linear unit (ReLU), sigmoid, tanh, or variants. The resultant points are placed in the same coordinated space as the activation space in the input image and are called feature maps. Each filter produces a feature map. The next layer is for the subsampling process, which aims to reduce the size of feature maps to reduce the number of parameters and make the algorithm faster. The common subsampling technics are Max-pooling and Average-pooling. As the name suggests, unit (ReLU), sigmoid, tanh, or variants.

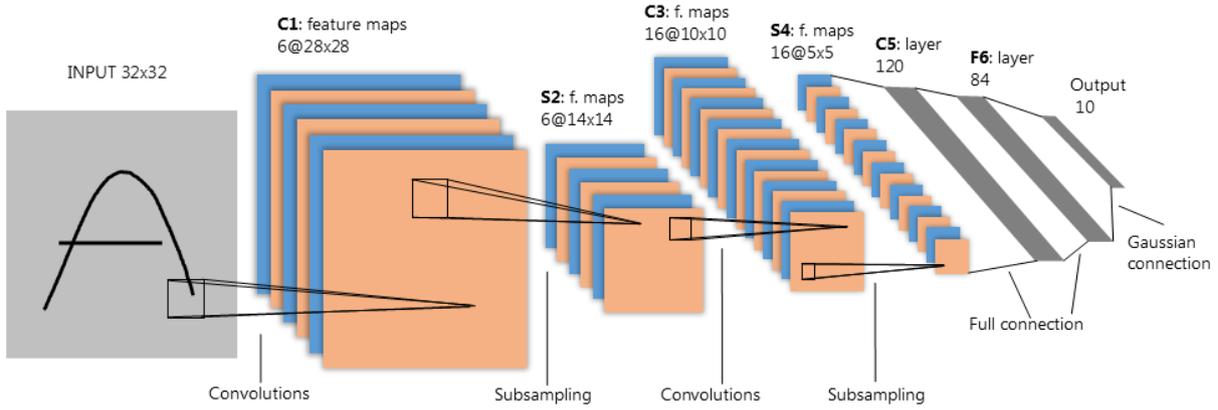

Figure 1 LeNet-5 Architecture. Source: Adapted from [1]

The resultant points are placed in the same coordinated space as the activation space in the input image and are called feature maps. Each filter produces a feature map. The next layer is for the subsampling process, which aims to reduce the size of feature maps to reduce the number of parameters and make the algorithm faster. The common subsampling technics are Max-pooling and Average-pooling. As the name suggests, from a window of pixels from the feature map, only the maximum value is selected or the average of pixels from that window, respectively. The pooling window size is a hyperparameter. It is often chosen as 2x2 for small resolution inputs such as the digits database 'MNIST.' The layers are staked while the model layers go deep. The last convolution layer is usually followed by fully connected layers and a classification or clustering method depending on the objective of the task.

Filter, also known as weight or kernel, transforms the input space into another space or output. Regardless of the architecture types and objective functions, filters are the core element in the whole "learning" process. The size of the filter is often considered a hyperparameter. In MLP, it is evident that the filter matrix depends on the number of neurons of current and previous layers. The CNN typically has images as inputs, and the algorithm breaks them into smaller regions called receptive fields (overlapped or non-overlapped), which are 2-dimensional (2D). Due to the nature of the convolution operation, the receptive fields and filter(s) must be of the same size and shape (2D), specifically in supervised learning. The logic is the same as the fundamental algebraic concept

$$Outcome(Y) = Weight(W) * input(X) + bias(b) \qquad (1)$$

For a nonlinear neural network, an activation function is implemented over the outputs.

$$Outcome(Y) = Activation\ [Weight(W) * input(X) + bias(b)] \qquad (2)$$

The purpose of the activation function is to control the flow of information depending on its behavior. Nonlinear activation functions add non-linearity, often claimed as biologically plausible or implausible. A deeper model up to a certain level is believed necessary for learning enough features depending on data complexity. Theoretically, multiple cascaded linear models can be replaced by a single linear layer, but the logic is disputed for computer vision architecture. The most widely used structures are cascaded layers with nonlinear activation functions [7]. Bias term is also noted to have a crucial role in firing neurons. Filters have some key parameters:

1. filter size and shape
2. Number of filters per layer
3. Learning algorithm

The research on these critical parameters focuses on the learning technic, classifying them into supervised, semi-supervised, self-supervised, and unsupervised categories. Moreover, The NN models in pattern recognition started from a multi-layer perceptron (MLP) (also known as fully connected NN). Later, convolution NN took over the computer vision field with promising results.

The contribution of this study can be listed as follows

- *Though many studies have been conducted on different architectures focusing on specific applications, learning types, or arbitrary groups, filter initialization and designing are often least discussed and studied. The filter initialization impacts the algorithm's convergence more than the actual learning algorithm. The current study comprises the most promising computer vision architectures regarding filter designing arguments and discussion. The parameters of filters, the filters' size and numbers of filters in different layers in DCNN architecture, training algorithm, training sample size, layers, and many other factors are aimed to get understood with some recent research in supervised and unsupervised learning approaches in DCNN.*

The rest of the paper is organized as follows. Section 2 focuses on filter initialization, its importance, and various technics. The main part of the paper is Section 3, and 4 comprises the arguments for specific selections of filter sizes and the number of filters throughout the network in prominent supervised and unsupervised methods, respectively. Section 5 summarizes the paper's findings, and the conclusion is presented in section 6.

## 2 | FILTER INITIALIZATION

Filter initialization is a crucial factor that impacts the final accuracy more than the learning algorithm in some cases in feed-forward networks (FNNs) [8]. Smaller initial weights result in smaller gradients which slows down the training. Larger initial weights can cause saturation or instability at activation; hence optimal weight initialization is crucial for preventing the output of any layers from getting exploded or vanishing through activation and is believed to be a critical factor for speed and ability to converge [1][9]. When CNN was proposed, it was common to set the initial weight as Gaussian noise with 0 mean and standard deviation of 0.01. Over the years, other initialization technics have also been proposed to prevent exploding, vanishing gradients, and dead neurons [10]. The optimum filter initialization is also seen as an open question with its relation to training sample labels, architecture, objective function, and types of outcomes of the algorithm [8]. Three scenarios are commonly used to initialize the weights.

### 2.1 | Random initialization

In random initialization, the values are randomly initialized near 0 and usually follow a normal or uniform distribution. The issue with random initialization is inconsistency. If the input values are too small, the convolution process would create a significant difference with epochs and result in different outcomes. Small values imply slow learning, prone to local minima, and possible vanishing gradient issue. On the other hand, large initialization values could saturate the neuron's outputs. They also could create exploding gradient issues, which would result in oscillation around the optimum target or instability condition [1][8][10].

For deep networks, some alternate versions have been proposed. LeCun [1] proposed uniform distribution between -2.4/Fi and 2.4/Fi, where Fi is the number of the input nodes. The reason is to have a similar initial standard deviation of the weighted sum to be in the same range for all nodes and ensure they fall in a specific operating region of the sigmoidal function. In contrast, it establishes a relationship with certain activation functions. Also, it is only feasible to apply when all connections sharing similar weights belong to nodes with identical Fi [1]. The general term for variance could be k/n, where k is a constant and depends on the activation function, while n is either the number of input nodes to the weight tensor or the number of input and output nodes of the weight tensor. Other widely used examples of random initialization are Xavier/Glorot and He initialization [8]. They use a normal distribution with mean zero and variance 1/n and 2/n, respectively. However, in some cases, uniform distributions are also used. Xavier is simple and sets the activations' variance the same across every layer. However, it is not applicable when the activation function is non-differentiable. He initialization overcomes this limitation and is widely used with the ReLU activation function, which is non-differentiable at zero [11]. LeCun, Xavier, and He do not eliminate the issue of vanishing or exploding gradients but mitigate the problem to a better extent [8].

### 2.2 | Zero or constant initialization

In this type of initialization, all the weights are set to either zero or a constant value (usually 1). As all the weights are the same, the activation also results in the same value, which results in symmetric hidden layers. In the supervised approach, the derivative of the loss function remains similar for all the nodes in a filter. Distance-based clustering methods would not benefit from this initialization since the constant value would only mimic the input values [8][10].

### 2.3 | Pre-trained initialization

Compared to the above initializations, pre-trained initialization is a reasonably new approach. There are two types of pre-trained weights: Transfer Learning, in which trained weights from any pre-trained model are borrowed and used as the initial state before starting new learning for the current method. Knowledge transfer accelerates the learning and generalization process. Earhan et al. [12] proved the claim by conducting comprehensive experiments on existing algorithms with pre-trained weights over the traditional approach. Pre-trained weights are also used in the student-teacher method, where a large network is trained extensively, and the optimized weights are transferred to comparable lightweight architecture for lighter applications [13].

Another initialization approach is to use a specific method that learns pattern distribution from the data and uses it as the filter's initialization. The principal component analysis (PCA) was used to initialize a self-organizing map as a filter, and faster convergence was observed [14]. K-means-based pre-initialization was experimented with and compared with random initialization. The K-means initialization uses an orthonormal matrix to rotate centroids symmetrically for an optimal solution through iteration. The generated filters are shown in Figure 2. Some filters learned with random initialization are prone to be noisy, as highlighted in red. K-means-based initialization would reduce noisy filters and result in faster convergence. It was also noted that any clustering method could produce a similar outcome [15].

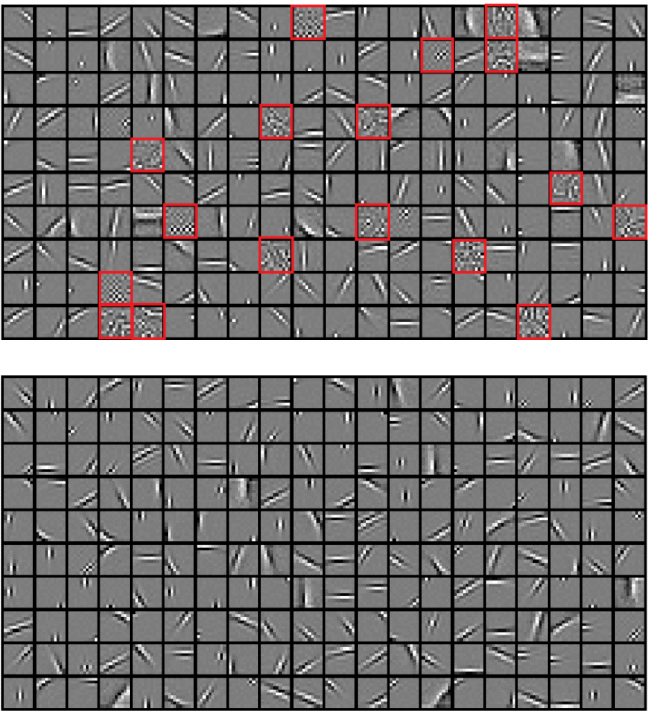

Figure 2 filters learned after random initialization (Top) and after using K-means-based initialization (Bottom) [15]

## 3 | FILTER DESIGN IN SUPERVISED LEARNING

In CNN, the convolution operation is the key to learning, and as per its nature, the filters must be of the shape of 2D to convolve with 2D image patches. Each filter aims to learn a specific feature that varies depending on the model's objective function. Different levels of correlation can be examined by utilizing different filter sizes since convolutional operations consider the vicinity (locality) of input pixels, yet the optimum size of a filter is an open question. Researchers exploited spatial filters to improve performance and investigated the relationship between a spatial filter and network learning. Also, the number of filters can vary, controlling the spectral resolution of the feature maps. We have tried to understand the deciding factors through various arguments over several significant architectures. This section discusses filters in terms of filter size (spatial resolution and spectral resolution) throughout layers and the number of filters in layers for promising supervised approaches. Filter size is noted as l x w x d @ z where l x w represents spatial resolution, d shows spectral resolution, and z is the number of filters.

### 3.1 | LeNet-5

There are many supervised learning algorithms for image classification which are widely used in the majority of real-world applications. However, CNN has been the backbone for most of the promising concepts. The LeNet-5 is the first significant structure for CNN. LeNet-5 has three convolutional layers followed by two fully connected layers. The filters' size in the first convolution layer was chosen 5x5@6 to have fewer connections. In the subsequent layers, they were selected as 5x5@16 and 5x5@120, "as small as" possible, to constrain the architecture size [1].

### 3.2 | AlexNet

AlexNet is another influencing architecture that reignited the research community's interest in CNN [2]. It has five convolutional layers. The filters' sizes from the first to fifth layers are 11x11x3@96, 5x5x48@256, 3x3x256@384, 3x3x192@384, and 3x3x192@256, respectively. The bigger filter size in the initial layers is selected to have a balanced number of sliding operations of convolution on the larger spatial resolution of the input image; 224x224. However, no supporting argument is given for selecting specific sizes over the subsequent layers. The network's depth and width and other parallelly linked modules contribute to the number of parameters and the complexity of the network. A smaller dataset can cause overfitting, and larger datasets may find a smaller number of parameters inadequate to get "learned" [2][5]. There are 60 million parameters in AlexNet, and the dataset was small. As a solution, data augmentation was implemented. The filters of earlier layers are considered extremely high and low pass filters as the mid frequencies have very little coverage. In a study, the filter size for the first layer was decreased to 7x7 from 11x11. The convolution stride was changed to 2 from 4 to reduce the aliasing artifacts in the second layer visualization. This stride change increased the parameters, but the extraction of features was improved [16].

### 3.3 | Visual Geometry Group (VGG)

In VGG (Figure 3), the depth was increased by adding more convolutional filters, and faster convergence was achieved using smaller filters. The larger filter size (e.g., 5x5, 7x7) in the previous architectures tends to increase the number of calculations, leaving the process computationally expensive. As a proposed solution, 3x3-sized filters with one stride were chosen for the whole architecture. With this filter size, two and three staked convolution layers provide an effective receptive field, the same as a single layer with a filter size of 5x5 and 7x7, respectively, without spatial pooling. A nonlinear convolutional layer with filter size 1x1 was also implemented. The advantage of this arrangement of layers with chosen size with implicit regularization is to gain faster convergence with fewer epochs. The filters were also pre-initialized [3]. In a similar work to VGG, another architecture named GoogleNet was proposed with smaller convolutional filters (sized 3x3 in addition to 1x1 and 5x5). GoogleNet has a few similarities with the Inception module and has 22 convolutional layers. Smaller filters were implemented in the earlier layers and larger ones in the later layers. It was claimed that smaller filters would reduce computation in the first few layers [3].

### 3.4 | Inception

A novel study was performed on the shape of the filters and implemented as the Inception module. In practice, doubling the filter bank sizes will result in four times as many parameters and four times more computationally expensive. Even though small filters are applied, the overall computation increase with the spatial resolution of the input images. The study aimed to reduce the dimension of the input representation [17]. It was hypothesized that removing highly correlated adjacent units results in less loss of information without any profound adverse effect. It was also added that optimal performance could be obtained by balancing the number of filters in layers and the depth of the model. Whether the width or length increases, performance increases to a certain extent, but the optimal way is to increase the width and depth parallelly. Unlike earlier studies, it was observed that filters size less than 5x5 in earlier layers do not capture the correlation between the signals and the activation of the units. Specifically, the 3x3-sized filter might suffer from a lack of expressiveness. However, this limitation was overcome using an Inception module where two stacked 3x3 convolutional layers replace each 5x5 convolution without any pooling layer.

For the Inception module, asymmetric filters are hypothesized to perform faster over symmetrical size. The proposed asymmetric (nx1) filters were claimed to have produced similar receptive fields when two convolution layers using filter sizes 3x1 and 1x3 to that of 3x3-sized filters. It was observed that replacing 3x3 filters with 2x2 reduced the computations by only 11%, while any nxn filters replaced by two convolutional layers with filter sizes such as nx1 and 1xn could save 33% of computations. This reduction in computation even increases with n. The Inception module increased efficiency by factoring the process into a series of operations having independent tackling cross-channel and spatial correlation [17]. The asymmetric filters have not been popular and have not been investigated.

### 3.5 | Residual neural network (ResNet)

Another prevalent architecture of deep neural networks is ResNet (Figure 3). It has an almost similar concept to VGG with some modifications in the basic design: (i) changing the stride of 2 for convolution operation, which splits the size of the feature map compared to its original size, and (ii) doubling the number of filters and keep their size same for all the layers to have feature maps of the same size to preserve the time complexity of the layer. Though filter size was kept as 3x3 as VGG, the model was built with more layers ranging from 34, 50, to 200 [4]. In another ResNet variant, wide generalized residual blocks were proposed; however, they do not contribute to the filter design [18]. While VGG was proposed as a balancing approach for increasing the width and length of the model in synchronization, ResNet and WideResnet were studied with varying lengths and widths of the network, respectively. The width of the network is directly proposed to increase the filter sizes and the number of filters, which was hypothesized to increase the representational power. WideResnet was also focused on making the algorithm more hardware friendly and argued that wide layers are more computationally effective than smaller filters as parallel processing can be faster on large tensors. However, optimum performance depends on the ratio of the number of ResNet blocks and the widening factor, and it is a hyperparameter. It was also observed that increasing the number of filters per layer is sufficient for learning, and performance can be improved as long as having adequate depth [19]. However, the optimum number is still believed to be data-dependent.

### 3.6 | Xception

Factoring the convolution into multiple branches was proposed in the Xception model [20]. It was claimed to be advantageous both on channels and in space. The architecture is a linear stack of depth-wise separable convolutional layers with residual connections. However, no discussion was found on its effect on filter size and numbers, as the study aimed to study the connection among the layers. Filter designing factors related to connections among layers have not been studied extensively, and no concrete information exists [20].

### 3.7 | DenseNet

In DenseNet, the first layer has filters of size 7x7 and then is reduced to 3x3 and 1x1 after each dense block. The structure can be observed as similar to residual-ResNet. However, a new concept of taking feature maps was proposed. Traditionally the feature maps are generated by the convolution and fed to the next layer, while in DenseNet, the convolution output is fed to the Dense block. In a Dense block, there are multiple layers, and the last layer directly connects with all the previous layers. It was claimed to have a better information flow and a more straightforward training process [21].

Numerous implementations of the above algorithms exist, e.g., Barlow Twins [22][23], Knowledge Distillations [24], learning via ignoring [25], Fusion of Quality maps [26], Dual-Stream Convolution-Transformer segmentation framework [27], DIFFNet [28], Few-shot learning [29], Vision Transformer [30] and many hybrid models. Most of the semi-supervised learning technics use the smaller portion of the labeled dataset to get trained and then tuned on the larger unlabelled dataset or vice-versa. Such hybrid algorithms use Alexnet, ResNet, VGG, and similar variants, as the backbone architecture. In such cases, the contribution regarding filter design is rarely observed.

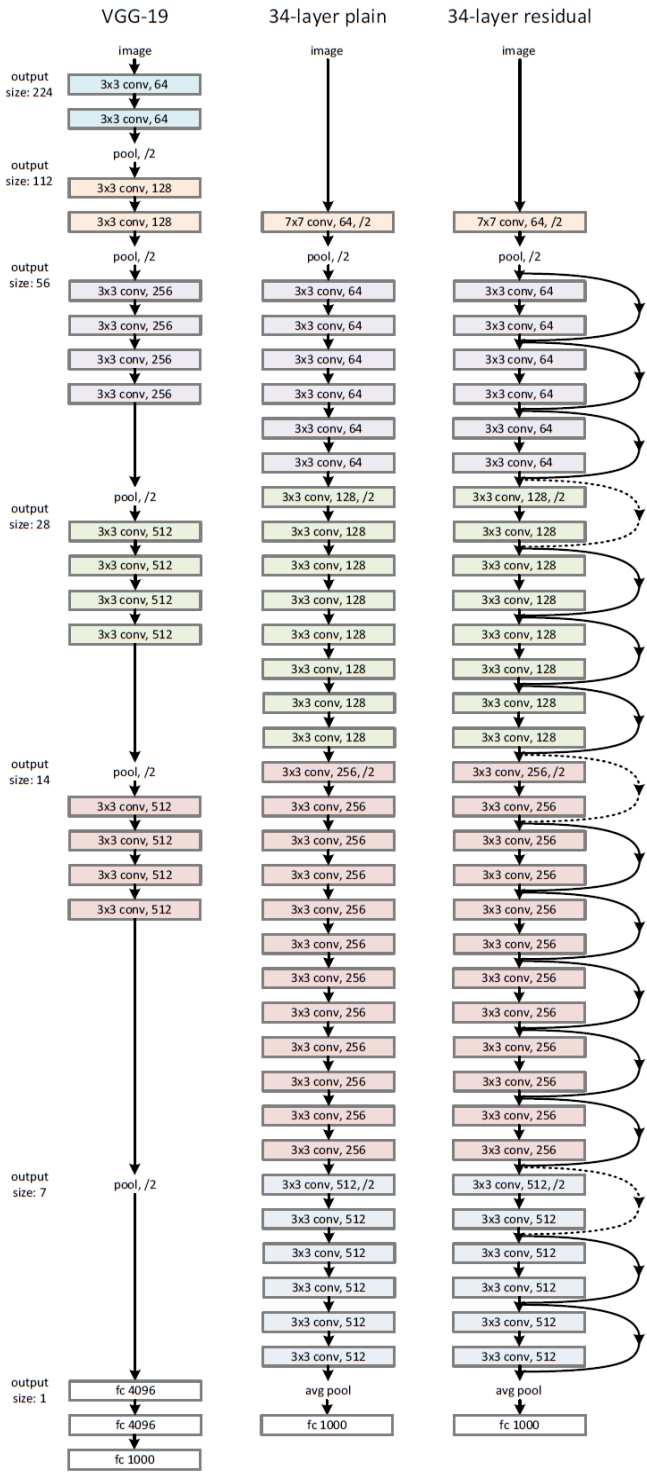

Figure 3 VGG-19 [3] (left), ResNet-34(Middle) and ResNet-34 with residual blocks [4]

The filter size and depth of the network are directly proportional to trainable parameters and the input data size [5]. The experiments from LeNet-1 to LeNet-4 concluded that the convolution network's size is directly related to the training set for the optimum use of datasets [8]. For LeNet-5, it was noted that larger architecture did not perform any better with increased complexity and had longer training time. A similar result was observed for ResNet-152, DenseNet-121, and DenseNet-169-201 versus AlexNet-8 [31]. The smaller network trained with a smaller dataset tends to perform weaker due to insufficient training [1] For AlexNet, it was noted that the architecture suffered from under-training for 1000 classes due to the small training dataset size, and data augmentation was used to meet the requirement [2]. To reduce the number of weights, deeper models like DensNet-100+, ResNet-50+, and similar variants have been modified, such as connection dropout and skipping layers. However, the filter sizes in deeper models are the same as discussed in traditional supervised models and offer no variety. To summarize the filter size adopted in the discussed architectures, we may conclude that the widely used filter sizes are 1x1, 3x3, 5x5, and 7x7. The larger sizes are chosen in initial layers to capture key spatial regions, while later sections use comparably smaller sizes for higher-level features. From the learning perspective, they mostly use backpropagation with Adam and SGD optimizer. Few semi-supervised methods also claimed to use various unsupervised techniques for fine-tuning.

## 4 | FILTER DESIGN IN UNSUPERVISED LEARNING

The drawback of supervised learning is the requirement of domain-specific large and labeled datasets, which is not often the case in the real world. Unsupervised learning does not require a labeled dataset; sometimes, a smaller dataset is enough. One of the most significant advantages of unsupervised learning is backpropagation-free training, which is faster and lighter. However, unsupervised training has two significant challenges: Effective learning and labeling. In the supervised methods, classes have labels according to the labeled datasets, while unsupervised methods develop clusters (based on the underlying structure of the data). However, the developed clusters need sensible labels, which is a different challenge. Many algorithms are proposed as self-labeling methods (self-supervised or semi-supervised) that primarily use a small set of labeled data. Labeling is not the scope of the current study, and the following discussion focuses on unsupervised learning. Unsupervised approaches in computer vision are mainly probability-based and distance-based learning. Probability-based methods use likelihood probability scores, while distance-based methods use "distance" as a primary deciding factor. The distance-based clustering technics are widely adopted. A smaller distance between objects indicates more' similarity,' and they might belong to the same or nearby clusters. There are three significant distance measuring methods: Cosine similarity, Manhattan distance, and Euclidean distance. Cosine is based on the angle between vectors (dot product); the smaller the angle, the more similar the vectors are, regardless of their magnitude. Euclidean distance is the L2 norm of the distance between the vectors in a Euclidean space, while Manhattan is the L1 norm.

In computer vision, unsupervised approaches are used for feature learning and clustering. The mixture of more than two architectures would result in a hybrid structure, i.e., when clustering is used after a typical DCNN section or for feature extraction, followed by a support vector machine (SVM) or any other classification method. In the later cases, the clustering method does not contribute to actual feature extraction, as clusters are formed based on features learned by DCNN. Authentic unsupervised learning happens when the algorithms are implemented in their typical nature to detect the object or used as filters in typical CNN architecture or a mixture of two or three models. In their traditional form, unsupervised methods do not use convolution operations for feature extraction. Such methods would go a little diverse from the focused study; however, studying them is mandatory to understand their implementations as convolutional layers.

4.1 | Probabilistic Methods
4.1.1 | Restricted Boltzmann Machine (RBM)
The restricted Boltzmann Machine (RBM) is an energy-based nonlinear probabilistic model. An RBM has a single visible and hidden layer having bidirected and symmetrical connections between the hidden and visible layers. It is typically used for dimensionality reduction, feature learning, modeling, and more. The RBM as a learning method has been studied extensively in the recent decade [32][33][34][35].

The RBM gets trained by the maximum likelihood rule using the contrastive divergence learning procedure. The pixels or inputs are the visible units as they are observable. At the same time, the feature detectors can be described as hidden units that capture the dependencies. The total number of weights = the number of nodes in the hidden layer times the number of nodes in the visible layer. Each pair between the hidden and visible layer is provided a probability via an energy function normalized to the sum of all possible pairs. The goal is to minimize the energy by adjusting the weights and biases for a particular image and raising the energy for other images.

The number of hidden layers is increased to deepen the model for better representations. For example, deep belief Net (DBN) is built by stacked RBM units. Furthermore, changing the depth, connections, and directions of communications have resulted in many other architectures like the deep Boltzmann machine (DBM), the shape Boltzmann machine (SBM), and the deep energy model (DEM) [32][34][36]. Two-layered sparse RBM was implemented over a raw natural image dataset. The filters learned in the first layer were observed to be local, oriented, and edge-detecting. The first and second layers had 400 and 200 filters, respectively. A snapshot of some of the learned filters from the first layer is shown in Figure 4 [33].

Convolutional RBM (CRBM) was proposed to tackle the curse of dimensionality to alleviate the computational cost [37]. Another reason for adopting the CRBM is to keep the 2D structure of images, which is not preserved in traditional RBM-DBN. In CRBM, the weights between hidden and visible layers are shared over the input image. In a study, multiple CRBM stacked one over another with max-pooling would result in convolutional DBN (CDBN). The filter size was chosen 10x10 for both layers, with the first layer having 24 groups (bases) while the second layer has 100. The visualization of filters for both layers is shown in Figure 5. It was observed that features learned in deeper layers were high-level and more specific to a particular object category [37]. Based on the CRBM, a deconvolutional network was proposed with only the decoder section, unlike traditional Encoder-decoder architecture [38]. For rich feature learning, Sparse decomposition using convex $l_1$ sparsity term was applied over the whole input image over the small image patches. The cost function was minimized over filters and feature maps during the learning. The process follows layer-wise starting from the first layer. The filters were 9, 45, and 150 for the first three layers. Notably, the filters from the first layer ensemble the outcome of the Gabor filters. The filter size was not discussed; however, it is mentioned to have a small size to avoid slow implementation of the proposed method. For larger filter size, FFT was recommended over spatial convolution for a faster process [38].

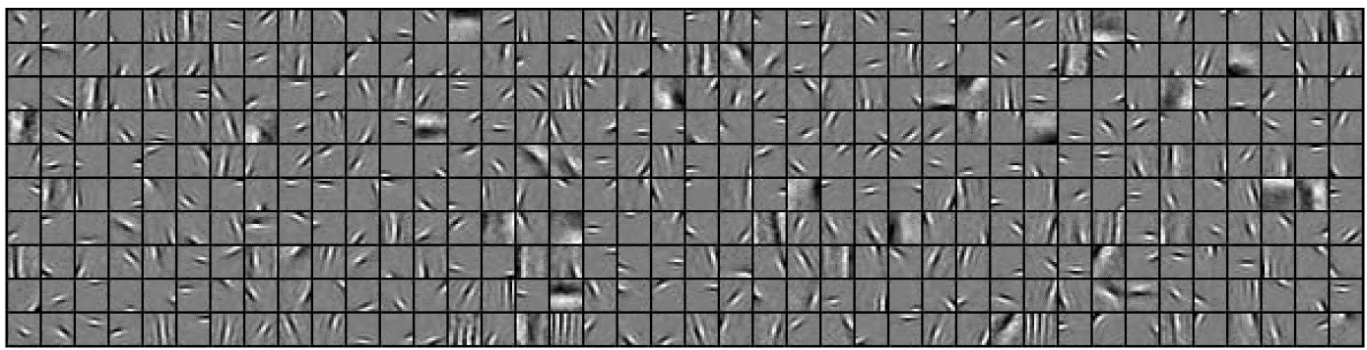

Figure 4 The 400 filters in the first layer of sparse RBM from the van Hateren raw image dataset [33]

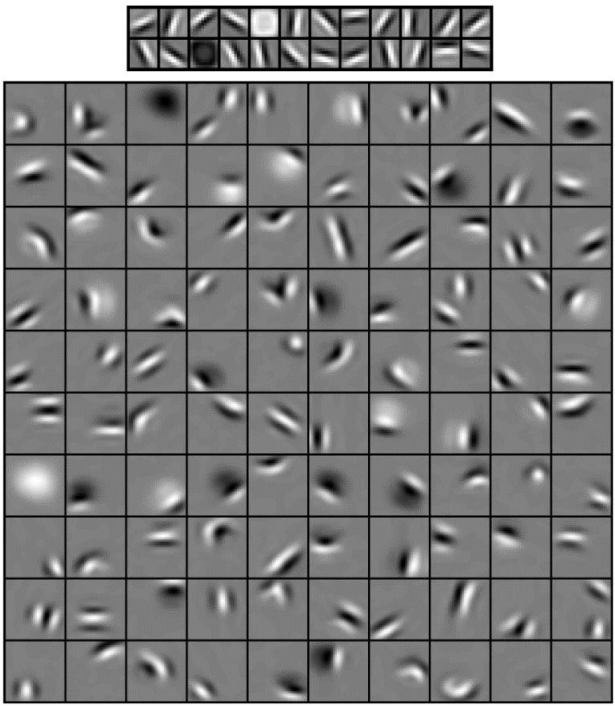

Figure 5 filter learned on first(top) and second layer(Bottom) bases. Second-layer filters can be viewed as a linear combination of the first-layer basis. [37]

4.1.2 | Autoencoder (AE)

The traditional autoencoder (AE) is a three-layered, fully connected network. Unlike MLP, AE is divided into two sections: encoder and decoder. The middle section has the lowest dimensions and is called the bottleneck layer. AE with a single hidden layer with linear activation can be similar to the principal component analysis (PCA). PCA is a widely used dimensionality reduction technic that gives orthogonal basis functions. It is a linear method and provides excellent results; however, it ignores features with low variances; it can omit some information that may be vital for feature learning. Furthermore, most datasets are nonlinear; autoencoder is adopted as a nonlinear dimensionality reduction method to overcome the limitation of PCA.

The deep autoencoder has more than three layers with nonlinear activation functions. The typical structure of the AE can be explained in Figure 6. Initially, the RBM architecture was used as the building block of deep AE. This method reshapes the image to 1D before processing it further. This original version of AE is more like a stacked RBM with reduced width of layers as it goes deep into the encoding section. As mentioned earlier, the AE has encoding and decoding sections; generally, the filters in the decoding section are the transpose of the encoding filters. This arrangement aims to preserve the symmetry of the architecture and reduce the parameters to be trained [37]. AE is an unsupervised method that creates a latent representation of the input by trying to regenerate the same input by the decoder and minimize the error. Stochastic gradient descent (SGD) is widely used to optimize the filters [39].

With the advancement of CNN, the AE got convolution layers instead of RBM as the building block. However, the 1D version of the image does not preserve information of 2D or 3D image layout. The issue is resolved in a modified architecture named convolutional AE (CAE) [40]. The deep CAE, commonly known as CAE, is used for object classification as a semi-supervised and unsupervised technic based on the method with which it is merged. In both cases, the compressed bottleneck core is calculated and used as the input to any classification model like MLP with SoftMax classifier, SVM [41], or to the clustering method like Gaussian Mixture Model (GMM) [40], K-means [37] [42], and self-organizing map (SOM) [43]. Many combinations of multiple methods are widely studied and used as self-supervised, semi-supervised or unsupervised architectures [37]. It was noted that decreasing the number of nodes with depth would be a trade-off between dimensionality reduction and feature extraction. The rate of which solely depends on the objective function of the application. When AE is used to regenerate an image's missing sections, the bottleneck layer's specific dimension is assumed unnecessary [44][45].

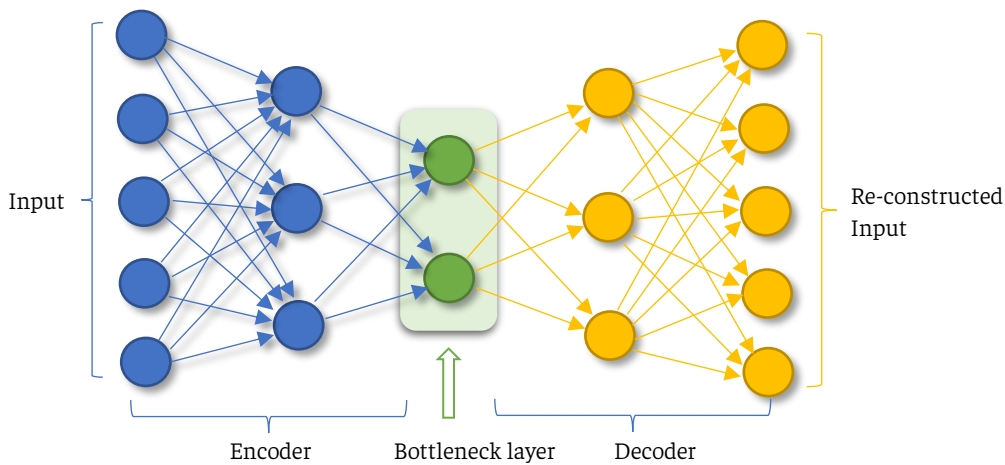

Figure 6 Typical autoencoder architecture. The first half is the encoding section resulting in a bottleneck layer (middle section), followed by a decoding section which is mostly the inverse of the encoding section.

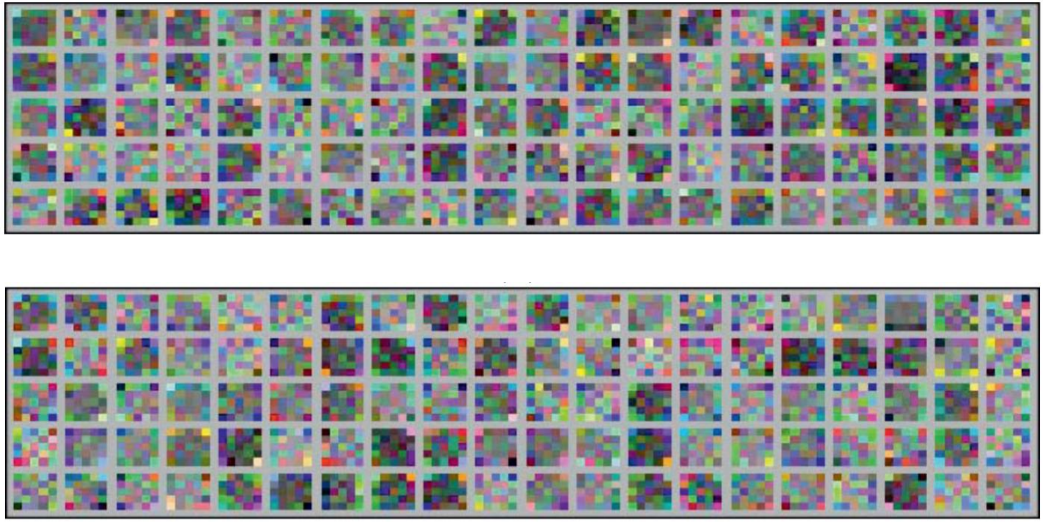

Figure 7 Filter learned in the first layer on the CIFAR-10 dataset by traditional CAE (no pooling and 0% noise) and denoising CAE (no pooling and 50% Gaussian noise) [46]

On the other side, the bottleneck layer is the essence of the data when AE is used for a dimensionality reduction-based feature extractor. The optimum size of the bottleneck layer is seen as dataset-dependent and a hyperparameter. For example, it was chosen 256D for ORL and Yale Dataset [37], 60D for MNIST, 100D for Fashion-MNIST, 160D for the COIL20 dataset [47], and 160D for the Astronomical dataset [39]. The encoder section in CAE is usually derived from successful CNN models like Alexnet [44], ResNet [47] or random convolution layers [42] that inherit the properties of a typical CNN architecture. The filter size remains 3x3, 5x5, and 7x7, the same as traditional CNN, while the number of filters and depth remain hyperparameters [40].

The variational auto-encoder (VAE) offers better organization of the latent space than typical AE by adding a loss function with a regularization term [48]. The standard AE is prone to learn only the identity function during the training if hidden layers are increased. Denoising autoencoder (DA) counters this problem by adding noise to the input, forcing AE to denoise and reconstruct it [49]. The DA offers excellent denoise capabilities with no distinct difference in filters, as shown in Figure 7. In the experiment, the filter size was set 7x7, 5x5, and 3x3 over MNIST and CIFAR-10 datasets with max-pooling layers of 2x2 [46].

4.1.3 | Convolutional Sparse coding

Sparse coding is a group of biologically plausible algorithms that leans the basis function to capture the high-level features from the unlabeled dataset [50]. In Sparse coding, a sparse vector is computed through the linear operation with a learned dictionary matrix for the optimum input reconstruction. Sparse coding was compared with batch alternatives and claimed to be an efficient feature learner over a large dataset [51]. However, the traditional sparse coding is computationally expensive [38] as it is performed over the whole image, and the representations are often redundant as inferences are performed at the patch level.

Koray et al. [52] applied sparse coding as a convolution filter bank learning mechanism (dictionary). The learned filters predict quasi-sparse features. The study compared the patch-based sparse coding model with the convolutional sparse coding model, and generated filters are shown in Figures 8 and 9. The filters generated by convolutional sparse coding reduced redundancy between feature vectors at nearby locations and increased overall efficiency.

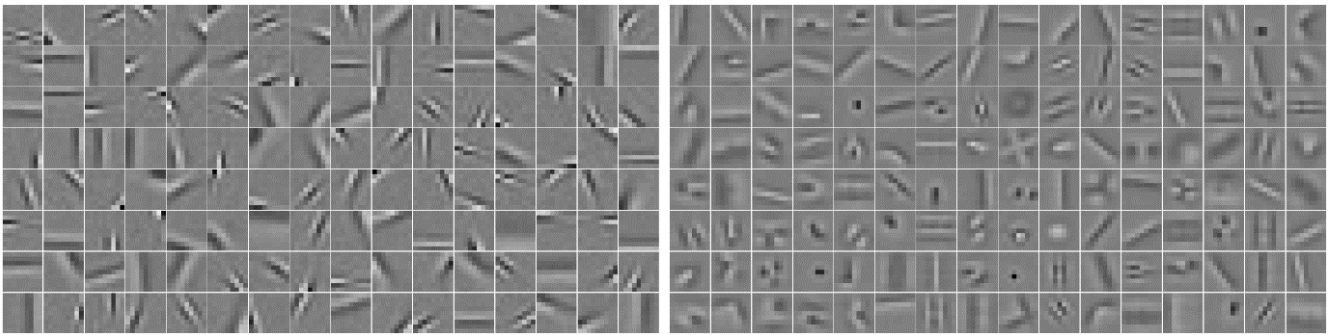

Figure 8 A dictionary of filters learned in the first layer with patch-based sparse coding model (left) and convolutional sparse coding model (right). Notably, the convolution model generates richer filters than the patch-based model. [52]

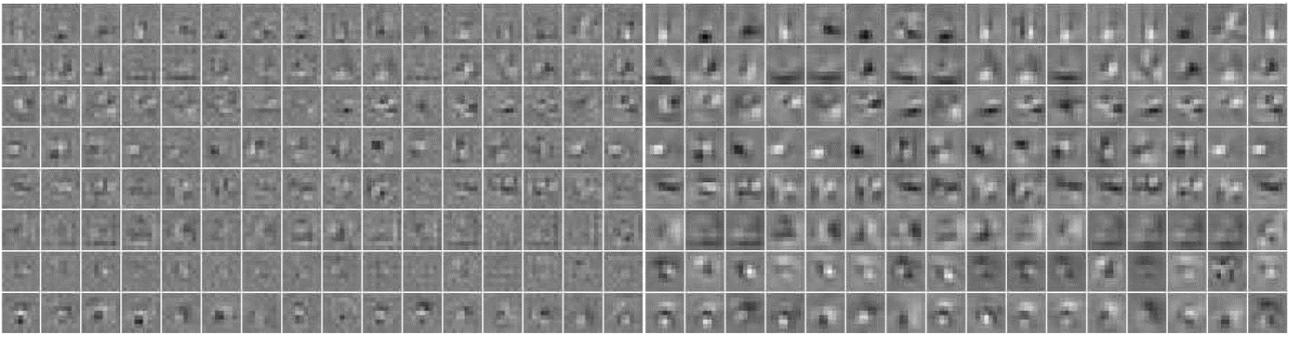

Figure 9 Second-stage filters of convolutional sparse coding [52]

The size of the convolutional filter bank was calculated using a formulation-based convolution rather than the input-dependent dictionary size. A formulation-based convolution was used over the traditional convolution approach to reduce the complexity. The filter size was selected as 9x9x64 and 9x9x256 for the first and second layers, respectively. However, no concrete arguments are made for particular sizes and numbers. It was claimed to have a better number of learned filters over convolutional RBM and traditional sparse coding [52].

4.2 | Clustering

Clustering methods aim at grouping the data points that possess "likeliness" measured by similarity measurement. Even though the groups have labels or not, grouping the "similar" data points is the core concept of clustering algorithms. The "similarity" can be defined in many ways. Mainly distance-based or probabilistic partitional-based methods are used for image clustering. The distance-based methods mostly use Euclidian or cosine measures, and probabilistic methods use probability scores in decision-making. The widely used performance criteria for cluster assignment are intra-cluster compactness and inter-cluster separability. The goal is to minimize cluster compactness and maximize the distance between clusters [37]. Compared to supervised methods, clustering methods require very little domain knowledge. In semi-supervised architecture, mainly supervised algorithms are used for feature learning, followed by clustering methods for grouping the objects (as an alternative to the supervised classification method). However, it was noted that training convolutional filters using clustering techniques can be promising and obtain general-purpose visual features [53]. A few distance-based methods have been applied to feature extraction as filters, and such studies are briefly discussed here.

4.2.1 | K-means

K-means is a commonly adopted clustering algorithm due to its simplicity. The fundamental concept is to find the centroids that minimize the distance between the points and the nearest centroids of the clusters in Euclidean space. The number of clusters (K) is the main hyperparameter needed to be defined initially. K-means as a learning module (feature Learning) can lead to excellent results; however, changes are required depending on the variety of datasets and objective functions [54].

Adam and Andrew [54] used K-means to obtain a dictionary of linear filters. The filter size was chosen 6x6 over input images of 32x32 and convolved with the stride of 1. The number of filters (K1, K2, and K3) for the three layers was chosen as 1600, 3200, and 3200, respectively. The experiment was focused more on selecting local receptive fields, and no discussion was found on determining the number of clusters over the three layers [55]. In a different approach, when the K-means clustering was applied to images for feature learning, the data points were considered pixels or image patches and centroids as the filters. The patch size was set the same as the number of centroids. In the experiment, the patch size is a hyperparameter and was selected 16x16; hence, the centroids dictionary was set to 256 [54]. The patches were selected randomly from the input, and the number of selections was around 10000. It was treated as a hyperparameter; however, random patch selection can be avoided now with large datasets available. K-means centroids efficiently detect low-frequency edges but perform poorly in the recognition task. As a solution, the whitening of images was performed before the filters' training, as whitening tends to generate more orthogonal centroids [54].

A. Dundar et al. [56] compared classical K-means with convolutional K-means as feature learning filters. Figure 10 shows the filters learned via classical k-means and convolutional K-means. The filters (highlighted red boxes) in classical K-means are likely shifted versions of each other, creating many centroids with similar orientations and generating redundant feature maps [56]. The widely noted issue with classical K-means is a decrement in efficiency with increasing input dimensions. Even for small images, the patch size directly affects the quality of learning by K-means filters. The patch size beyond some point results in poor performance, and the optimum size (taken 6x6 or 8x8) remains a hyperparameter. The depth of the model is directly proportional to the number of trainable parameters. As a solution, a random selection of patches is observed as a widely accepted solution [53][55][56].

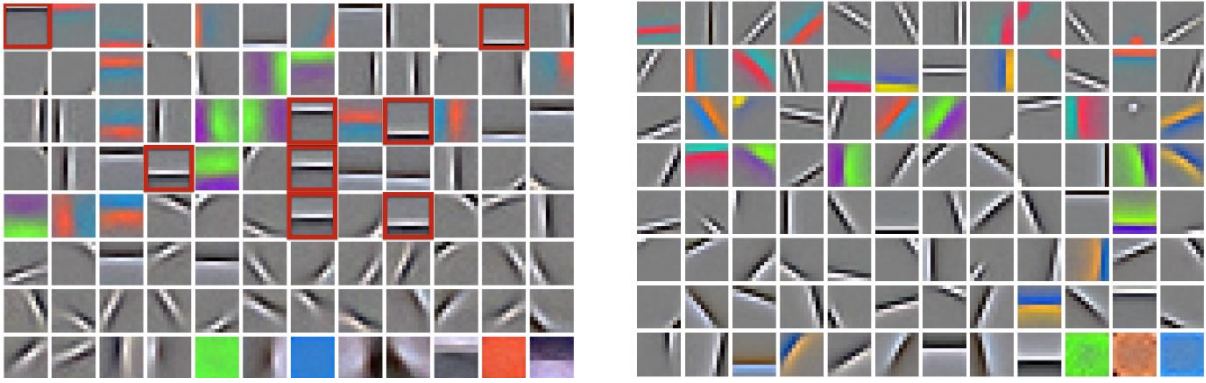

Figure 10 filter trained by K-means (left) and convolutional K-means (Right) over STL-10 dataset [56]

Convolutional K-means was claimed to have significantly reduced the redundancy in centroids compared to classical K-means. In convolutional K-means, the size of windows from the input images is two times bigger than the filter size and randomly selected. The convolution operation would be performed over the entire window with centroids of K-means filter to obtain a similarity matrix at each location. The patch with the most significant activation from the window can be considered the closest to the centroid, and its specific location can be assigned to the matching centroid [56]. The filter size and the number of filters were treated as hyperparameters for optimum performance. While keeping the filter size fixed at 11x11, the number of filters varied and increasing the number of filters resulted in better performance. Similarly, the number of filters was fixed at 96, with the filters' size varied, and the results were compared. In both cases, it was observed that the filters learned through convolutional K-means consistently outperformed those learned by classical K-means. However, the method is not fully unsupervised, as a backpropagation-trained weight matrix was used for learning the connections.

4.2.2 | Self-organizing Map (SOM)

Self-organizing map (SOM) is another widely used clustering method. Like K-means, SOM is also based on distance measurement; however, SOM can be interpreted as a constraint K-means method. Traditional SOM can learn and cluster similar attributes on a 2D map. The Euclidian distance is measured between all the neurons on the map and input data. The node on the map with minimum Euclidian distance is the Best matching unit (BMU). Unlike K-means, the neighborhood neurons of BMU are also updated; subsequently, the whole map represents the learning model. Though SOM is a very simple and powerful clustering technic, the major drawback is its shallow structure which may not be sufficient to learn efficient information-carrying features from big datasets, specifically images [57].

A relatively recent study proposed SOM-based multi-layer architecture named convolutional SOM (CSOM) [58]. The method has convolutional layers placed between SOM layers. The novelty was the convolutional-based feature learning by the SOM. The SOM map was used as a filter for learning, followed by a convolution layer with that learned filter. Figure 11 shows a snapshot of filters generated by SOM. The structure was tested with two types of pooling layers: Traditional max-pooling and SOM-based pooling. The latter approach performed better with the feature map with learned SOM maps (filter). The winner neurons were chosen using Euclidean distance. The SOM map size was chosen 8x8 for input images of size 256x256. However, no specific reason was mentioned for choosing a specific map size.

Alexander and Ayanava [59] proposed a biologically plausible architecture named Resilient self-organizing Tissue (ReST) that can be executed as a typical CNN. The continuous energy function of SOM was the core of the study. It was noted that traditional SOM suffers from optimum convergence state and model parameter values, while an energy function can provide a simple quality measure. With continuous energy function, Stochastic gradient descent (SGD) can be extended to SOM learning in deep learning. Unlike the traditional SOM, the learning rate over time was kept constant. The Map size KxK was treated as a hyperparameter for K ∈ {10, 15, 20, 30, 50} and chosen 10x10 for varying input batch size of NxN, where N ∈ {1, 5, 10, 20, 50, 100}. A larger map and batch size would significantly increase the training time.

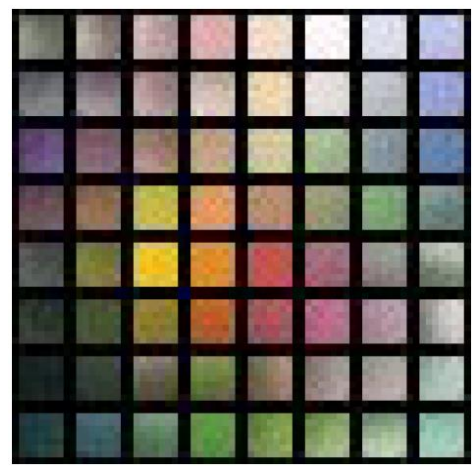

Figure 11 Filter learned by SOM in the convolutional SOM (CSOM) method [58]

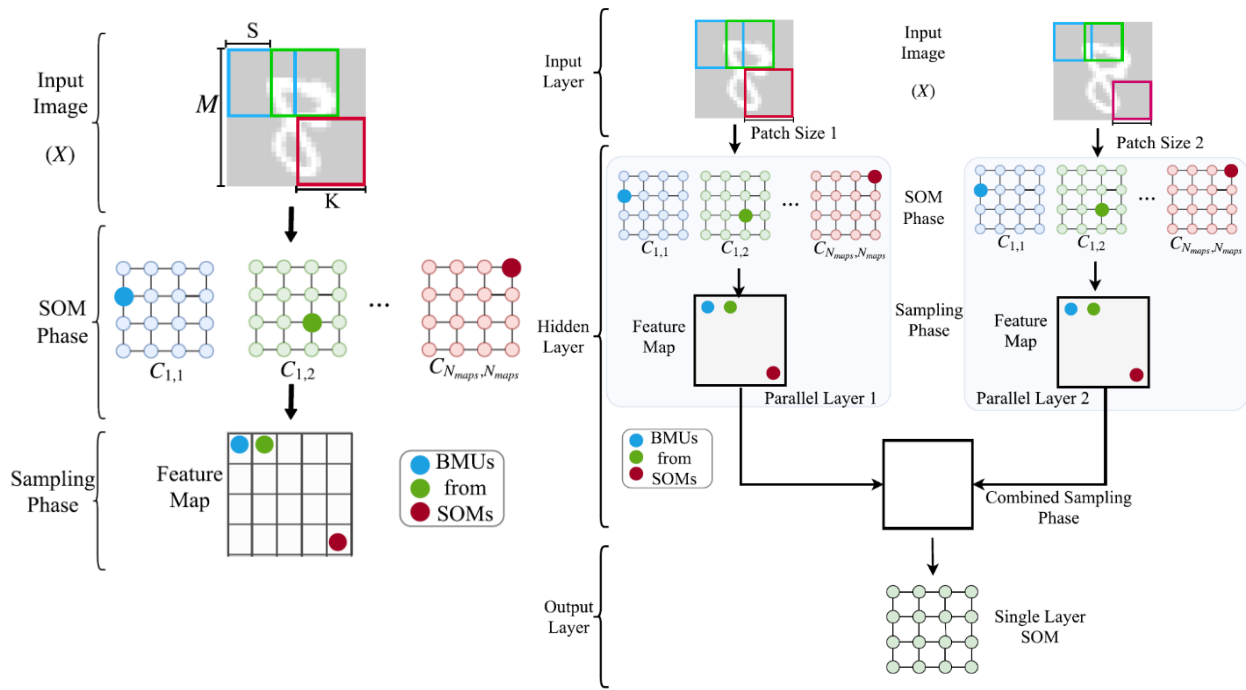

Figure 12 DSOM (left) and E-DSOM (Right). SOMs act as filters [57]

In a similar approach to CSOM, two more architectures were proposed, named Deep SOM (DSOM) and extended DSOM (E-DSOM) [60][57]. The block diagrams of both architectures are shown in Figure 12. In DSOM, each activation space gives a winner on a SOM Map (filter) during the SOM phase. The next layer is the sampling phase, in which the feature map is formed. Each node in the feature map is a BMU from an activation space and is stored in the relevant space of the activation space. The E-DSOM has multiple DSOM architectures running in parallel, and the feature maps are combined at the end. In two-layered architecture, the output layer gives a single SOM map and is used as an input to a classification method. The map size varied from 4-24 for the first layer and 14-16 for the second layer using MNIST, GSAD, and SP-HAR datasets. The E-DSOM outperformed DSOM with a classification accuracy of up to 15% with time-saving by 19%. The downside is the requirement for more computational power for parallel architecture.

When traditional SOM is used as a filter and convolved on the input image, it was observed that the SOM tries to fit the dataset best which may lead to poor performance. To overcome this issue, the Hebbian learning-based masking layer was multiplied with the input patches before convolution with SOM maps (filter) [61]. The two-layered architecture used 10x10x1 and 16x16x1 sized SOM maps for the first and second layers, respectively. In three-layered architecture, the layer-wise SOM map sizes were selected as 12x12x3, 14x14x3, and 16x16x3. On a trained SOM map, a three-layer MLP classifier was used. It was also observed that more than one filter is required to learn enough distinct features. Also, multiple smaller maps are better than fewer large maps [61]. Valued-SOM (VSOM) was proposed as an improvised version of DSOM with the introduction of the Lethe term to each output neuron. The purpose of this mechanism is to introduce a supervision mechanism for self-labeling to the clusters created by DSOM. However, the filter design was borrowed from the original DSOM [62]. Kosmas et al. [63] proposed Dendritic-S method, which uses SOM as feature extraction filters followed by a hit matrix for labeling. The cosine similarity was applied and compared with Euclidian distance in determining the BMU, and accuracy was improved by nearly 20% in the experiments.

SOM maps are known to have 1D and 2D maps, but a 3D SOM map as a filter was introduced in a model named deep convolutional self-organizing map (DCSOM) [14]. A total of 256 nodes were applied in various dimensions ranging from 1D to 6D. The 4D(4x4x4x4) was noted as the optimum map size balancing performance and complexity. The map dimensions higher than four resulted in overfitting. The input patch size of 5x5 was found optimal from 3x3 to 15x15. The other focus of the research was the radius of the neighborhood of the SOM map and the batch learning technic. The two-layered convolutional model was followed by a block-wise histogram for feature representation [14]. A computationally effective, Unsupervised-DSOM (UDSOM) method was proposed in which the SOM maps were passed through ReLU activation after the learning phase. The aim is to remove the neurons from the maps that never get activated or become BMU, which could lead to fewer connections. [64] The map size was chosen in the four-layered model as 10x10, 8x8, 6x6, and 4x4, respectively. The smaller-sized filters performed better for higher-level features. A faster version of UDSOM was proposed as G-UDSOM, which performed parallel processing of locating BMU for patches over different maps [65]. However, the UDSOM was used as the backbone architecture, and there is no modification to the filter sizes (map).

4.2.3 | Sub-space learning (SSL)

The implementation of sub-space learning in object identification has gained much attention recently. The core of those proposed architectures is principal component analysis (PCA) which comes under unsupervised learning and is mainly used for dimensionality reduction.

The subspace approximation and kernel augmentation (Saak) is an earlier proposed SSL-based algorithm [66] Saak is a one-pass feed-forward method that was proposed as a solution to the limitations of the older RECOS (REctified-COrrelations on a Sphere) method. The backpropagation and nonlinear activation functions ReLU in the RECOS method cause approximation and rectification losses, respectively [7]. The structural insight of the Saak method is shown in Figure 13. Filters are generated in Saak using the subspace approximation with second-order statistics and orthonormal eigenvectors of the covariance matrix. The filters are based on truncated Karhunen-Loeve transform (KLT) or PCA, which are the unit eigenvectors of the data covariance matrix. Such filters are automated and generated based on the dataset, and their size can be varied. Saak has non-overlapping convolutional operations with filters and patches of size 2x2. ReLU is a widely used activation function where the negative inputs are truncated to zero, resulting in rectification loss. In the Saak, all the kernels are augmented with their opposite counterpart. When the original kernels and their augmented parts pass through ReLU, the positive sides survive, which can be either original or augmented. This mechanism would result in no rectification loss. Saak is an entirely new methodology for better interpretability of deep networks. However, it has some limitations because of the increased computation implicated by kernel augmentation.

Jay Kuo at el. [67] proposed an improved version of Saak named Saab (Subspace approximation with adjusted Bias). Saab, a variant of PCA, has convolutional layers followed by MLP. Typically, the bias term varies for all layers. However, Saab was set as a constant equivalent to the most negative value of the input vector, making the nonlinear activation function redundant. The convolutional filters were obtained from the covariance matrix of bias-removed spatial-spectral cuboids and were chosen of size 5x5. The convolutional filters generated by PCA capture a large amount of energy, but the capacity decreases with more significant indices. The higher the cross entropy, the lower the discriminant power. After convolutional layers, the MLP was used for labeling, and parameters were calculated using linear least-squares regression (LSR). It was claimed as a novel approach to self-labeling [67].

Multiple Saab structures were used in cascade in PixelHop architecture. The purpose is to learn features from different neighborhood sizes (hop). Like Saab, earlier layers use a higher energy ratio, while later layers use a low energy ratio. Frobenius norms of the difference of the two-covariance matrix were used for faster filter convergence, which rapidly converges to zero. For the CIFAR-10 dataset, four cascaded Saab structures were implemented. The patch and filter sizes were 4x4, 4x4, 2x2, and 2x2 for all the PixelHop units starting from the first to the fourth structure [68]. Channel-wise Saab transform was applied in PixelHop as the building block, and the new architecture was named PixelHop++. The structural overview is shown in Figure 14. This technique reduced the filter size and memory requirement for filter computation [69].

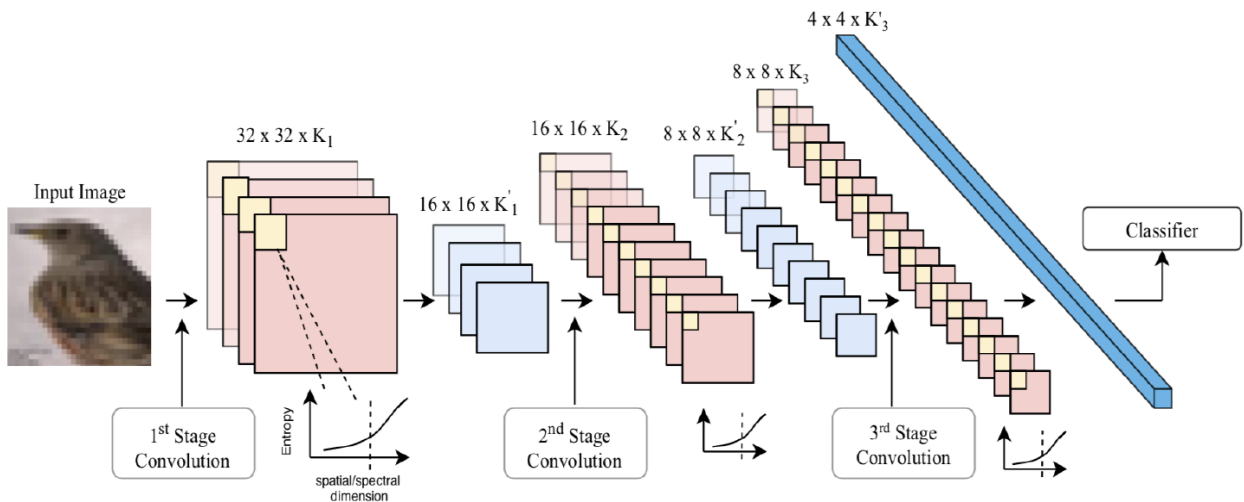

Figure 13 Three-stage Saak architecture [70]

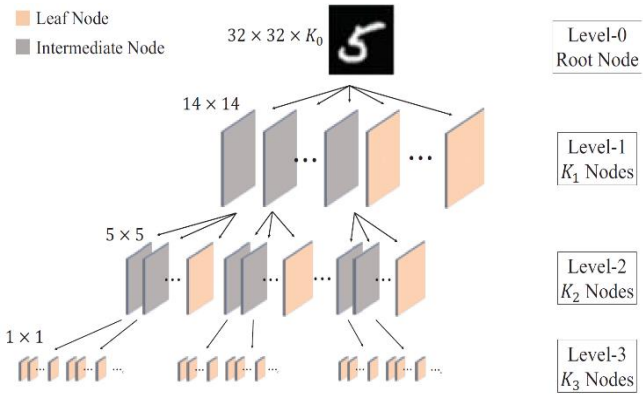

Figure 14 Channel-wise Saab transform – PixelHop++ [69]

For feature learning, a novel tree-decomposed method was proposed whose leaf node provides 1D features, and the features are sorted from lowest to highest by their cross-entropy [71]. The features of lower cross-entropy have higher discriminative power. The filter size was chosen 5x5, inherited from the channel-wise Saab architecture. The number of filters was also increased for convolution layers 6 &16, 16 &32, and 32 &64 for the first and second layers, respectively, and applied for MNIST, Fashion MNIST, and CIFAR-10 datasets. An energy-based hyperparameter Threshold (T) was added to determine the model's size. The threshold was an excellent parameter for trading minimal efficiency with fewer connections. For example, for MNIST, the change of T from 0.005 to 0.00001, the efficiency decreased by just 0.51%, but parameters were reduced by four times. To make the PixelHop++ lighter, Enhanced-PixelHop (E-PixelHop) was proposed, in which multiple PixelHop++ units were used as the building block. To reduce the number of parameters, only two components were considered in the PCA analysis [71]. However, there was no change in the size and number of filters from the original PixelHop++ architecture. A few other models named FaceHop, VoxelHop, and DefakeHop were proposed. However, they use Saab architecture with minor modifications, but the basic design of filters is the same.

4.3 | Contour Detection architectures

Before the popularity of CNN, many feature extraction methods were proposed. One of the promising concepts is Independent Component Analysis (ICA). ICA resembles the receptive fields of simple cells in the visual cortex. ICA on natural images produces phase and frequency-sensitive decorrelated filters, which resemble oriented Gabor functions (Figure 15). ICA is different from PCA and cannot be calculated analytically. ICA requires the minimum mutual information among the output vectors and is achieved by choosing a nonlinear activation function as the cumulative density function of the underlying independent components. ICA filters were observed to have more sparsely distributed outputs on natural images [72].

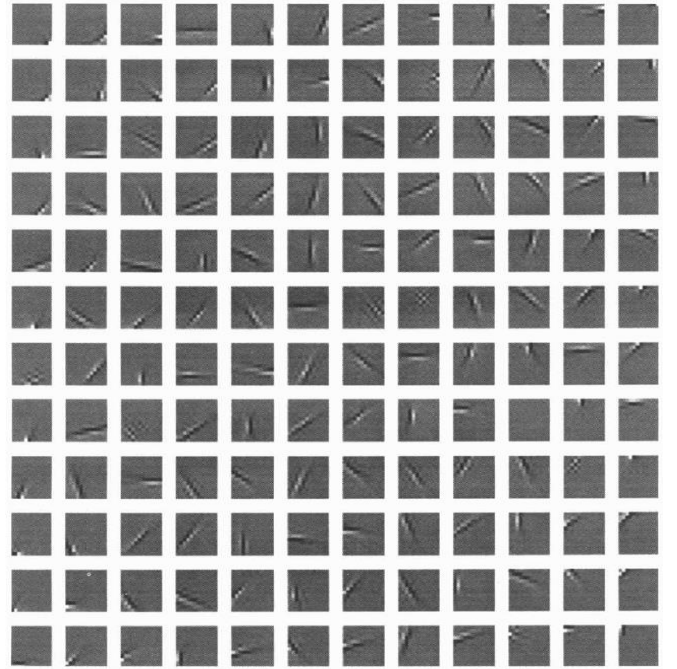

Figure 15 filters trained through ZCA-whitened natural images, which are visually the same as the ICA filters [72]

In 2007, Thomas [73] used a biologically plausible method, a feed-forward path of object detection. It used the Gabor Function [74] to tune the filter size. It is claimed that the Gabor function is more biologically inclined and has been proven to be an effective model for simple cortical cell receptive fields. The model is divided into simple and complex units. The filter sizes were calculated from 7x7 to 37x37@16 for simple units and 8x8 for complex units.

Sobel and Prewitt filters are popular gradient-based methods for Edge detection in the frequency domain. Mathilde et al. [53] proposed 'deep-cluster,' a novel K-means-based clustering approach for large-scale, end-to-end training of convolution layers. The paper was focused on optimization based on K-means, but the K-means clustering was not a part of the main convolutional architecture. AlexNet was arbitrarily selected as the backbone architecture and can be replaced with any similar architecture. The first layer used the Sobel filter as a feature extractor. The Vertical and Horizontal weight matrices for Sobel filters are:

$$W_{ver} = \begin{bmatrix} -1 & 0 & 1 \\ -2 & 0 & 2 \\ -1 & 0 & 1 \end{bmatrix} \text{ and } W_{Hor} = \begin{bmatrix} 1 & 2 & 1 \\ 0 & 0 & 0 \\ -1 & -2 & -1 \end{bmatrix}$$

Most filters cannot get trained over raw images with colors, which is the primary motivation for applying the Sobel filter. The Sobel filter is pre-defined, and the values are fixed for edge detection in the vertical and horizontal directions. As no learning happens for the Sobel filter, it remains independent of further training. The Prewitt filters share a similar property to Sobel and could generate a similar result.

# 5 | SUMMARY

In the late 20th century, handpicked features were extensively studied. Restricted Boltzmann machine and subsequent auto-encoder have been applied directly for feature learning. The PCA, ICA, ZCA, Gabor filter and many statistical methods were applied and observed as promising concepts. After the LeNet-5, backpropagation-trained convolutional NN attracted much attention from the research community, and hybrid structures started to be studied. The lack of computational facilities and rich databases has left many algorithms abandoned for decades and almost forgotten.

Since the promising result of AlexNet in 2012 in the ImageNet competition, hundreds of supervised architectures have been proposed in various sizes and shapes in the last decade. As discussed earlier, they enjoy being different from a structural point of view; however, they use backpropagation as the primary training mechanism with mostly Adam, SGD, and RMSprop optimizer. The gradient calculation is computationally expensive, and gradients also have the issue of getting vanished or exploding. BP also suffers from weight transport problem, which is not believed to be happening in the human brain and is considered biologically implausible. As the model goes deep to achieve the complex objective function, the interpretability becomes vague due to nonlinear activation functions and a large number of nodes in hidden layers. The lack of interpretability leaves a minimal opportunity for significant change, and as a result, the supervised models are almost at the saturation point in terms of improvement. Another reason is the lack of large labeled datasets, which are essential for training but very costly in terms of time, and energy to obtain in the real world. Such models are data-driven; hence, implementation in different fields requires field-specific rich datasets, which is not an optimum solution to the universal problem of true Artificial Intelligence.

However, the concept of convolution is not a dead end; it is proven that convolutional NN is undoubtedly promising over traditional MLP for feature learning on images. However, the limitation lies in the learning mechanism. Various unsupervised architectures have been applied to counter the issue, ranging from probabilistic models to distance-based methods. The advantages of such models are being less complex, faster, and trainable even on a small unlabelled dataset, which is very convenient for broad applications. However, the downside is the efficiency; there is much room for improvement. Another issue with unsupervised learning is labeling the clusters due to the unlabelled training datasets. Self-labeling is also an open question, and many self-supervised and semi-supervised methods have also been proposed as a solution. Few-shot learning and one-shot learning have also given promising results in labeling clusters. For core learning, K-means clustering has been extensively studied and has provided superior efficiency among other clustering and probabilistic methods. However, K-means does not follow the topology of the patches during training. Robustness to the noisy dataset is a crucial requirement in real-world applications, and K-means is prone to noise. To counter these limitations, self-organizing maps (SOM) have been applied. Thanks to neighborhood learning characteristics, SOM preserves the topology of the input data. By that, the distance between nodes in the SOM map reflects those in the high-dimensional input space. However, K-means does a similar thing, but the visualization becomes tricky as the K-means cluster centers are not in a convenient 2D shape.

The SOM has been studied extensively in feature learning with the original form and as convolving filters. However, these models have been shallower compared to supervised models. The deeper models have been proven to be necessary for considerable learning. In DSOM, EDSOM, and UDSOM, the nodes which have not been BMUs have been dropped either in the selection process or via ReLU. In D-CSNN, the SOM weight vector was traditionally trained, and BMU was calculated using the Hebbian-trained mask layer. The approach was unique and promising. However, it was left with no subsequent experiment. Convolutional sparse-coding also generates good features but lacks sharpness. It is one of the least studied algorithms.

The dimensionality reduction technic was also tried to use as a feature extractor. One of the most studied technics is PCA, in which the orthonormal basis function derived from the covariance matrix can give promising results. However, PCA is a linear technic and underperforms with the non-linearity of real-world datasets. PCA also ignores the low variance, and the importance of features with low variance is still not known for the image dataset. Kernel PCA is an extension of PCA that links nonlinear relations between cluster nodes and high dimensional input; however, it has not been studied. As a solution, auto-encoder, a nonlinear unsupervised dimensionality reduction technic, has been used extensively. Though AE and CAE are more like traditional MLP and CNN-based architecture with decreasing size of hidden layers, they offer no variety for filter design.

As mentioned earlier, the lack of interpretability is choking the improvement of CNN-based models. The CNN was tried to explain as a multi-layer RECOS (REctified-COrrelations on a Sphere) model. It was a supervised approach trained with BP. The Saak and Saab provide insightful interpretations of CNN and novel learning concepts. In both of them, the filters for convolution layers are calculated by PCA. In Saab, selecting a sufficiently large bias term eliminates the requirement of nonlinear activation; however, it may increase the value of weights, and the whole purpose of standardization may vanish. K-means for a pseudo-label generator in the FC layer via linear regression is a novel concept. In Saab, Saab layers were not trained with FC in an end-to-end training fashion. End-to-end learning has produced promising results as a deep cluster and may improve the filters having a feedback loop. It is vital to investigate the labeling of the clusters with minimum use of labeled datasets. Some research has been done on labeling using online algorithms to prevent any human intervention in self-learning. It is not the scope of this paper, but it is very tricky for an algorithm to decide which label to assign if not even a single labeled sample is given. At last, the effect of batch size and epochs have not been noted in this paper as almost negligible studies have been found on them and remain an open question.

# 6 | CONCLUSION AND FUTURE SCOPES

This paper provides a detailed review of arguments and factors affecting designing the filters for the most promising architectures based on the convolutional neural network. The paper introduces the convolutional neural network, its layers, and signal flow, followed by a brief overview of hyperparameters like filter size, number of filters, activation function, and more.

The primary purpose of this paper is to shed light on the factors and supporting arguments made in promising studies for designing filters. The review starts with the filter initialization and its importance and specifies different types of the same. Each of them is briefly explained with strong and weak points. Due to the different nature of learning, the filter designing study is divided into supervised and unsupervised learning groups. The filter designing parameters are discussed in detail for promising supervised methods like AlexNet, ResNet, VGG, and similar variants with subsequent versions. The relevance of these parameters on input data, objective functions, application types, computational power, and other parameters are noted and critically compared. We have surveyed and reviewed the studies on unsupervised methods like AE, K-means, SOM, and SSL with the same objective, and the arguments are concluded. Having lack of mathematical backing, filter designing is mainly summarized as a data-dependent process. Deep learning is still in its infancy, and having open questions like optimum filter designing is a tremendous opportunity for the current algorithms.

# 7 | REFERENCES